\documentclass{article}

\usepackage[utf8]{inputenc}

\usepackage{subfig}
\usepackage{caption} 
\usepackage{xcolor}
\usepackage{booktabs}
\usepackage{amsthm}
\usepackage{listings}
\usepackage{inconsolata}
\usepackage{tikz}
\usepackage{pgfplots}
\usepackage{amsmath}
\usepackage{microtype}
\usepackage{algorithm}
\usepackage{amssymb}
\usepackage[titletoc]{appendix}
 \usepackage{cancel}
 \usepackage{multirow}
\usepackage{subfig}

\theoremstyle{definition}
\newtheorem{definition}{Definition}

\lstnewenvironment{myalgorithm}[1][] 
{
    \lstset{ 
        basicstyle=\normalfont\ttfamily,
        showspaces=false,               
        showstringspaces=false,         
        mathescape=true,
        numbers=left,
        escapeinside={*}{*},
        columns=flexible,
        numbersep=2pt,        
        keywordstyle=\bfseries,
        keywords={,and, return, not, def, in, if, elif, else, for, foreach, while, continue,break,}
        numbers=left,
        xleftmargin=6pt
        #1 
    }
}
{}

\usepackage{pgfkeys}
    \def\addlegendimage{\scriptsize\csname pgfplots@addlegendimage\endcsname}
\usepackage{comment}

\usepackage{ijcai26}

\usepackage{times}
\usepackage{soul}
\usepackage{url}
\usepackage[hidelinks]{hyperref}
\usepackage[utf8]{inputenc}
\usepackage[font=small]{caption}
\usepackage{graphicx}
\usepackage{amsmath}
\usepackage{amsthm}
\usepackage{booktabs}
\usepackage{algorithm}
\usepackage{algorithmic}
\usepackage[switch]{lineno}

\title{Learning Logical Rules Using Minimum Message Length}

\author{
Ruben Sharma$^1$
\and
Sebastijan Duman\v{c}i\'{c}$^2$\and
Ross D. King$^{1,3}$\And
Andrew Cropper$^4$\\
\affiliations
$^1$University of Cambridge, United Kingdom\\
$^2$Delft University of Technology, the Netherlands\\
$^3$Chalmers University of Technology, Sweden\\
$^4$University of Helsinki, Finland\\
\emails
bars2@cam.ac.uk,
s.dumancic@tudelft.nl,
rk663@cam.ac.uk,
andrew.cropper@helsinki.fi
}

\begin{document}

\maketitle
\begin{abstract}
Unifying probabilistic and logical learning is a key challenge in AI. We introduce a Bayesian inductive logic programming approach that learns minimum message length hypotheses from noisy data. Our approach balances hypothesis complexity and data fit through priors, which favour more general programs, and a likelihood, which favours accurate programs. Our experiments on several domains, including game playing and drug design, show that our method significantly outperforms previous methods, notably those that learn minimum description length programs. Our results also show that our approach is data-efficient and insensitive to example balance, including the ability to learn from exclusively positive examples.
\end{abstract}

\section{Introduction}
Inductive logic programming (ILP) is a form of machine learning.
The goal is to induce a hypothesis (a set of logical rules) that generalises examples and background knowledge (BK) \cite{mugg:ilp,ilpintro}.

As with all forms of machine learning, ILP handles noisy data.
A popular approach is to search for a minimal description length (MDL) hypothesis \cite{cabalar2007minimal,hocquette2024learning}.
MDL is a formalisation of Occam's razor  \cite{mdl,rissanen1998stochastic,rissanen1987stochastic}.
The goal is to encode data as a binary string as concisely as possible. 

Minimum message length (MML) \cite{wallace2005statistical}
is another formalisation of Occam's razor.
The goal is to find the most concise explanation of data, where an explanation is a two-part message consisting of a hypothesis (the first part) followed by the data given that the hypothesis is true (the second part). 

MDL and MML are closely related, but differ in philosophy, mathematical grounding, and practical usage \cite{baxter1994mdl,makalic2022introduction}. 
We expand on the differences in Section 2, but briefly summarise them now.
MML is explicitly Bayesian and uses priors over hypotheses and integrates them into the message length. 
MDL avoids priors and often treats hypotheses as tools for encoding data rather than probabilistic entities. 
In MML, the message is two-part, first asserting the hypothesis, then the data given that the hypothesis is true. 
In MDL, the message may be one-part and often does not encode the hypothesis itself.
MML and MDL also handle continuous parameters differently \cite{baxter1994mdl,grunwald2007minimum}.

In this paper, we use the MML principle to learn hypotheses from noisy examples. 
Specifically, we design an MML cost function for learning hypotheses expressed as probabilistic logic programs \cite{raedt:problog}.
The cost of a hypothesis is the length of the message if we were to communicate the examples using this hypothesis. 

Our main claim is that our MML method outperforms an existing MDL method \cite{hocquette2024learning}, shown to achieve state-of-the-art performance when learning logic programs from noisy examples. 
We call this previous score \emph{C-MDL}.

Our MML method has three major advantages over C-MDL. 
First, the cost is the exact message length, resulting in more precise comparison between hypotheses. 
The increased precision enables us to determine when there is sufficient data to learn a more complex hypothesis, allowing us to generalise more reliably from sparse data.
Second, our method uses a structured likelihood function that penalises over-generalisation, which enables learning from unbalanced, or one-class data. 
Third, MML uses prior probabilities which allows us to favour more general hypotheses. Favouring more general hypotheses prevents overfitting, as we do not wish to learn overly specific conclusions from small amounts of data

Specifically, we claim:
\begin{description}

\item[\textbf{Claim 1:}] MML performs similarly to C-MDL on noisy data and general tasks
\item[\textbf{Claim 2:}] MML outperforms C-MDL when learning from different example proportions, including exclusively positive or negative examples
\item[\textbf{Claim 3:}] MML is more data efficient than C-MDL
\item[\textbf{Claim 4:}] An explicit prior preference for more general programs improves the data efficiency of MML
\item[\textbf{Claim 5:}] A practical approximation to MML outperforms C-MDL when learning from unbalanced examples
\end{description}
Overall, our contributions are:
\begin{enumerate}
    \item We formulate the ILP problem in Bayesian and MML terms. We use the MML framework to derive a cost function for learning probabilistic logic programs, consistent with the Problog semantics \cite{raedt:problog}.
    \item We introduce a practical method to learn approximate MML programs from examples.
    \item We empirically show that the MML score outperforms C-MDL on predictive accuracy and matches its performance on noisy data. The method excels when learning from few or unbalanced examples.
\end{enumerate}

\section{Related Work}
\textbf{ILP.}
Most ILP approaches learn from noisy data \cite{foil,progol,tilde,aleph}.
For instance, \textsc{MaxSynth} 
\cite{hocquette2024learning} searches for programs that minimise the total of the training errors and literals in the program. 
In contrast to our approach, these approaches do not allow for the explicit declaration of prior beliefs regarding the quantity of noise in the examples, do not employ techniques to mitigate the effect of example balance in their learning, and do not return probabilistic logic programs.

\textbf{Positive only learning.} 
Several ILP approaches learn from exclusively positive examples \cite{de1993theory,quinlan1995induction,mooney1995induction}.
Most assume example completeness. 
One exception \cite{muggleton1996learning} also learns in a Bayesian framework, but is restricted to noiseless data, assigns groups of hypotheses the same prior probability based only on size, and employs a greedy search to find the hypothesis of maximum posterior probability.
We differ by handling noise, using a more sensitive prior tailored to each program, and utilising a SAT solver in our search procedure.

\textbf{MML and MDL}. 
MML and MDL have been applied to learn a wide range of models, including neural networks \cite{makalic2003mml,hinton1993keeping}, decision trees \cite{wallace1993coding}, automata \cite{wallace2005statistical}, and statistical distributions \cite{farr2002complexity}.
\citeauthor{conklin1994complexity} \shortcite{conklin1994complexity} and \citeauthor{deterministic-ILP-paper} \shortcite{deterministic-ILP-paper} both use MML to learn deterministic logic programs.
\citeauthor{stochastic-extension} \shortcite{stochastic-extension} uses MML to learn probabilistic programs. 
None of the cost functions in these works consider negative examples and do not use an MML-based procedure for estimation of continuous parameters. 
By contrast, our cost function considers negative examples and we use an MML-based procedure for estimation of continuous parameters. 
Moreover, we use predicate priors to express an explicit preference for more general programs and we use a constraint-based search procedure. 
We choose MML over MDL for its Bayesian framework and method of handling parameter uncertainty: encoding real-valued parameters to a finite precision. 
MDL usually uses approximations or penalised maximum likelihood (ML) estimators to estimate continuous parameters. MML's handling often yields less biased and consistent estimators that are often more accurate than ML \cite{baxter1994mdl,wallace2005statistical}.

\textbf{Probabilistic logic programming}. 
Previous work has focused on learning rule probabilities \cite{gutmann2011learning} or learning rules from Bayesian networks \cite{de2008probabilistic}. 
Probfoil \cite{probfoil} is built on the Problog semantics \cite{raedt:problog} and learns probabilistic logic programs, but is frequentist, employs a greedy search, and requires both positive and negative examples. 
Markov logic networks \cite{richardson2006markov} are first-order knowledge bases with a weight attached to each logical formula and consider the joint distribution of a set of predicates.  Our approach is Bayesian, uses a constraint solver in the search, can learn from exclusively positive examples, learns a single hypothesis, and employs MML as the hypothesis selection criterion.
\section{Problem Setting}
\subsection{Inductive Logic Programming}
We define an ILP input in the learning from entailment setting  \cite{luc:book}: 
\begin{definition}[\textbf{ILP input}]
\label{def:probin}
An ILP input is a tuple $(E, B, \mathcal{H})$ where $E=(E^+,E^-)$ is a pair of sets of ground atoms denoting positive ($E^+$) and negative ($E^-$) examples, $B$ is background knowledge, and $\mathcal{H}$ is a hypothesis space, i.e a set of possible hypotheses.
\end{definition}
\noindent
Background knowledge is a definite logic program.
The hypothesis space contains Problog \cite{raedt:problog} programs of a particular syntactic form:
\begin{definition}[\textbf{Probabilistic hypothesis}]\label{def:hypothesis}
Let $g$ be a Datalog program, 
$f$ be the predicate symbol in the head of the rules in $g$,
$a$ be the arity of $f$,
$\theta_+$ and $\theta_-$ be probabilities $>$ 0.5,
and $g_{\phi}$ be $g$ where every instance of the predicate symbol $f$ is replaced with the  predicate symbol $\phi$, 
where $\phi$ is a new predicate symbol of arity $a$.
A probabilistic hypothesis $h = g_{\phi} \cup \{r_+,r_-\}$, where $r_+$ and $r_-$ are: 
\begin{align*}
r_+ &= \ \ \ \theta_+ :: &&f(X_1,\dots,X_a) \leftarrow \phi(X_1,\dots,X_a)\\
r_- &= 1-\theta_- :: &&f(X_1,\dots,X_a) \leftarrow \text{not } \phi(X_1,\dots,X_a)
\end{align*}
\end{definition}
\noindent
The function $\text{det}(h) = g$ maps a probabilistic hypothesis $h$ to its deterministic counterpart $g$.
Informally, $h$ is a probabilistic wrapper around $\text{det}(h)$, and $h$ makes identical predictions to $\text{det}(h)$ but assigns probabilities to the predictions, governed by $\theta_+$ and $\theta_-$. 

We define a cost function:
\begin{definition}[\textbf{Cost function}]
\label{def:cost_function}
Given an ILP input $(E, B, \mathcal{H})$, a cost function $\mathcal{C}_{E,B}~:~\mathcal{H}~\to~\mathbb{R}_{>0}$ assigns a numerical cost to each hypothesis in $\mathcal{H}$.
\end{definition}

\noindent
We define an \emph{optimal} hypothesis:
\begin{definition}[\textbf{Optimal hypothesis}]
\label{def:opthyp}
Given an ILP input $(E, B, \mathcal{H})$ and a cost function $\mathcal{C}_{E,B}$, an optimal hypothesis is $h^* = \arg\min_{h \in \mathcal{H}} \; \mathcal{C}_{E,B}(h)$.
\end{definition}

\noindent 
Given a hypothesis $h$ and background knowledge $B$, a true positive is a positive example entailed by $det(h) \cup B$. 
A true negative is a negative example not entailed by $det(h) \cup B$. 
A false positive is a negative example entailed by $det(h) \cup B$. 
A false negative is a positive example not entailed by $det(h) \cup B$. 
For brevity, we omit $E$ and $B$ and denote the number of true positives, true negatives, false positives, and false negatives of hypothesis $h$ as $tp(h)$, $tn(h)$, $fp(h)$, and $fn(h)$ respectively.

We describe the C-MDL cost function used by \citeauthor{hocquette2024learning}
\shortcite{hocquette2024learning}

\begin{definition}[\textbf{C-MDL cost function}]
\label{def:MDL_cost}
Let $h$ be a hypothesis with $fp(h)$ false positives and $fn(h)$ false negatives on training examples, 
and $size(g)$ denote the number of literals in the definite program $g$.
The C-MDL cost of $h$ is $\text{C-MDL}_{E,B}(h) = size(\text{det}(h)) + fp(h) + fn(h)$.
\end{definition}

\noindent
We describe a notion of \emph{entailed atoms} and \emph{entailed examples} with respect to a hypothesis
\begin{definition}[\textbf{Entailed Examples}]
\label{def:entailed examples}
Given a hypothesis $h$ and an instance space $X$ containing all possible examples, we define a set of \emph{entailed atoms} as $\{x \in X \;|\; B \cup det(h) \models x\}$ and a set of \emph{non-entailed atoms} as $\{x \in X  \;|\; B \cup det(h) \not\models x\}$.
Given a set of examples $E\subseteq X$, we define a set of \emph{entailed examples} as $ \{e \in E  \;|\; B \cup det(h) \models e\}$ and a set of \emph{non-entailed examples} as $\{e \in E  \;|\; B \cup det(h) \not\models e\}$.
\end{definition}
\section{Minimum Message Length} 
Given a prior and a likelihood function, MML aims to find the most concise two-part explanation of examples $E$. 
An explanation consists of a hypothesis $h$ (first part) and the examples given that hypothesis is true $E|h$ (second part). 
We briefly summarise this communication format following \citeauthor{wallace1993coding} \shortcite{wallace1993coding}.

Given a hypothesis $h$, the examples $E$ can be optimally encoded as a binary message of length:  
\begin{equation}
    C(E|h) = -\log_2 P(E|h) \nonumber
    \label{eq: second part}
\end{equation}
where $P(E|h)$ is the probability of observing $E$ given that $h$ is true.
This message can be decoded by a receiver who already knows $h$.  
If the receiver does not know $h$, the sender must also send $h$, using a code agreed upon \textit{a priori}.  
The length of this message is:  
\begin{equation}
    C(h) = -\log_2 P(h)\nonumber
\end{equation}
The total message length required to send the hypothesis and the examples is therefore:
\begin{equation}
    C(h,E) = C(h) + C(E|h) \label{eq:full cost}
\end{equation}

\noindent
Equation~\ref{eq:full cost} is the MML cost function, which balances hypothesis complexity and data fit. 
A complex hypothesis must explain the data succinctly to be selected over a simple hypothesis that less succinctly explains the data.
An MML optimal hypothesis is one that has the smallest value of $C(h,E)$. 

Equation \ref{eq:full cost} is formed of a hypothesis cost $C(h)$ and an examples cost $C(E|h)$.
We describe each at a high-level. Full derivations for omitted equations may be found in the Appendix.
\subsection{Hypothesis Cost}
The cost of sending a hypothesis $h$ (definition \ref{def:hypothesis}) is:
\begin{equation}
    C(h) = 
    \text{syntax}(h) + 
    \text{theta}(h)+
    \text{coverage}(h)
    \label{eq:hyp length}
\end{equation}
We discuss each component in turn. 

\subsubsection{Syntax}
The cost of the syntax of $h$ is the sum of three subcosts\footnote{We omit sending the total number of literals in $h$, as its cost is constant for any hypothesis}:
\begin{equation}
    \text{syntax}(h) = \text{structure}(h) + \sum_{r\in\text{det}(h)} \big[\text{predicates}(r)+ \text{vars}(r)\big]
    \nonumber 
\end{equation}
\noindent 
We discuss each subcost in turn.

\paragraph{Structure.}
    The cost of the arrangement of literals into rules in $\text{det}(h)$, denoted $\text{structure}(h)$, is based on the total number of ways that the non-head literals can be arranged into rules, which is the integer partition function.  
    The sender specifies a single arrangement.
    For instance, if there are $3$ total non-head literals in $\text{det}(h)$, there are $3$ possible arrangements:
    \begin{verbatim}
 f(X):-*,*,*.  f(X):-*,*.  f(X):-*.
               f(X):-*.    f(X):-*.
                           f(X):-*.
    \end{verbatim}
    Specifying one arrangement from the above three requires $\log{3}$ bits. 
    Formally, let $\text{body}(r)$ return the literals in the body of rule $r$ and $\text{part}(n)$ denote the integer partition of $n$. 
    Then:
    \begin{equation}
        \text{structure}(h)=\log{\text{part}\left(\sum_{r\in \text{det}(h)}|\text{body}(r)|\right)}
        \label{eq:arrangement of literals into clauses}
    \end{equation}
    \noindent
    The cost is greater for hypotheses with more non-head literals.

\paragraph{Predicate Symbols.}
    The cost of the predicate symbols in a rule $r$, denoted $\text{predicates}(r)$, is based on the individual predicate generalities, defined as the relative occurrence of the predicate in the Herbrand Base defined over $\text{det}(h)\cup B$ (see definition \ref{def:probin}).     
    Formally, let $o_p$ be the number of atoms that contain the predicate symbol $p$ in a Herbrand Base $HB$ and $ord(r)$ denote the number of possible orderings of the predicates in the rule. For example, given $r = f(X,Y)\leftarrow a(X), b(Y).$ there are $ord(r)=2$ orderings: $[a/1, b/1]$ and $[b/1,a/1]$.    
    Let $\text{sym}(r)$ return the multiset of the predicate symbols in $r$. Then
    \begin{equation}
        \text{predicates}(r) =-\log{ord(r)} -\sum_{p\in \text{sym}(r)}{\log\frac{o_p}{|HB|}}
        \label{eq: predicate cost}
    \end{equation}
    More general (larger $o_p$) predicate symbols result in a shorter message length. We only consider non-recursive programs, so we omit the counts of head predicates in the computation of $HB$.
    
    \paragraph{Variables.}
    The cost of the variables in a rule, denoted $\text{vars}(r)$, is based on the number of possible variable arrangements in the rule.
    Formally, let $\text{var\_comb}(r)$ represent the number of possible variable combinations in rule $r$, given a set of predicate symbols and a maximum variable count. 
    Then
    \begin{equation}
        \text{vars}(r) = {\log{\text{var\_comb}(r)}}
        \label{eq: variable cost}
    \end{equation}

\subsubsection{Theta Values}
    The cost of the theta values $(\theta_+, \theta_-)$ is based on a standard MML code (MML87) for binomial values \cite{wallace2005statistical,wallace1987estimation}, denoted as $\text{theta}(h)$ (see Appendix). 
    The MML87 code is shorter for theta values that are similar to our prior beliefs on likely values of $(\theta_+, \theta_-)$.
    
\subsubsection{Coverage}
    We also send the number of entailed examples (definition \ref{def:entailed examples}) denoted $N_+$, and the number of non-entailed examples, denoted $N_-$.
    These numbers are necessary to decode the message as the two cases are governed by two separate probability models (rules $r_+$ and $r_-$, see definition \ref{def:hypothesis}). The receiver must know how many examples to decode using each probability model.
    If $h$ makes no errors, then $N_+=|E^+|$ and $N_-=|E^-|$. 
    The more errors made by $h$, the larger the difference between $N_+$ and $|E^+|$. We expect accurate $h$, so state larger differences with longer messages (see Appendix). 
\subsection{Examples cost}
The cost of sending the examples $E$ is:
\begin{equation}
    C(E|h) = \text{atoms}(E|h)+\text{labels}(E|h)
    \label{eq:ex length}
\end{equation}
The receiver knows the set of possible examples (the instance space), but not the specific training examples or their truth labels which are sent separately. We describe each next.
    \paragraph{Atoms.} 
    We use the terms entailed atoms, non-entailed atoms, entailed examples and non-entailed examples as defined in definition \ref{def:entailed examples}.
    We assume $N_+=tp(h)+fp(h)$ example atoms are chosen from the entailed atoms, and $N_-=tn(h)+fn(h)$ from the non-entailed atoms.
    Then the cost of sending the example atoms, denoted $\text{atoms}(E|h)$, is based on two things.
    First, the probability of choosing the entailed examples.
    Second, the probability of choosing the non-entailed examples. 
    Then $\text{atoms}(E|h)$ is the sum of the negative logarithm of the two probabilities.
    Formally, let $e(h)$ and $\neg e(h)$ denote the number of entailed atoms and non-entailed atoms respectively. 
    Let $h$ have $tp(h)$ 
    true positives, $fp(h)$ false positives, $tn(h)$ true negatives, and $fn(h)$ false negatives on $E$. Assuming all $E$ are equally likely to be chosen, $\text{atoms}(E|h)$ is:
    \begin{equation}
        \underbrace{-\log{\binom{|e(h)|}{tp(h)\!+\!fp(h)}^{-1}}}_{\text{ entailed by det}(h)} - \underbrace{\log{\binom{|\neg e(h)|}{tn(h)\!+\!fn(h)}^{-1}}}_{\text{not entailed by det}(h)}\nonumber
    \end{equation}
    More probable choices result in a shorter message.

\paragraph{Labels.}
 The cost of the example truth labels, denoted $\text{labels}(E|h)$, is determined by their probability given $h$ under Problog semantics. 
The probability of each example truth label is determined by either $r_+$ or $r_-$ (definition \ref{def:hypothesis}). 
The probability of all example truth labels is the product of the individual truth labels: the product of two Bernouilli distributions which describe the entailed and non-entailed examples respectively (definition \ref{def:entailed examples}). 
Then $\text{labels}(E|h)$ is the sum of the negative logarithms of the two probabilities.
\begin{equation}
    \underbrace{-\log{\left[\theta_+^{tp(h)}(1-\theta_+)^{fp(h)}\right]}}_{\text{ entailed by det}(h)}- \underbrace{\log{\left[\theta_-^{tn(h)}(1-\theta_-)^{fn(h)}\right]}}_{\text{not entailed by det}(h)} \nonumber
\end{equation}
The $\text{labels}(E|h)$ encoding favours hypotheses that misclassify fewer examples.
\section{MML Example}
Consider the task of learning an integer concept $f/1$. Assume we have the background knowledge:
$$
\begin{array}{l@{\hspace{.25cm}}l@{\hspace{.25cm}}l@{\hspace{.25cm}}l@{\hspace{.25cm}}l}
  prime(2). & prime(3). & prime(5). & prime(7). & \\
  even(0) &even(2).& even(4). &even(6).&  even(8).\\
 odd(1). & odd(3). & odd(5). & odd(7). & odd(9).
\end{array}
$$
Examples $E = E^+\cup E^-$, where:
\begin{align*}
         &E^+ = \{f(2), f(5)\}  &&E^- = \{f(3)\}
\end{align*}
Consider the hypothesis $h = \{g_{\phi}, r_+, r_-\}$ where:
\begin{alignat*}{2}
&g_{\phi} = \phi(X)\leftarrow prime(X). \\
&r_+= 0.88 :: f(X) \leftarrow \phi(X). \\
&r_-= 0.14 :: f(X) \leftarrow \text{not }\phi(X). 
\end{alignat*}
From definition \ref{def:hypothesis}, the deterministic counterpart of $h$ is:
\begin{equation}
    \text{det}(h) = f(X)\leftarrow prime(X). \nonumber
\end{equation}
\subsubsection{Determining the Theta Values} 
We calculate the $\theta_+, \theta_-$ values of $h$ from $\text{det}(h)$, applying the MML87 procedure to the results of $\text{det}(h)$ on training examples: $tp(h)=2$, $fp(h)=1$, $tn(h)=0$, $fn(h)=0$.
Assuming a beta prior with parameters $\alpha=10$, $\beta=1$ for both $\theta_+$ and $\theta_-$, MML87 gives $\theta_+ = 0.88$ and $\theta_- = 0.86$. Note $1-\theta_- = 0.14$. 
\subsection*{Hypothesis Cost}
We calculate the costs of $C(h)$ (equation \ref{eq:hyp length}).
\paragraph{Structure.} Given that there is only one non-head literal in $h$, there is only one arrangement of literals into rules and thus $\text{structure}(h)=\log{\text{part}(1)}=\log{1}=0$ bits (equation \ref{eq:arrangement of literals into clauses}).
\paragraph{Predicate symbols.} 
There is only one rule $r\in\text{det}(h)$ with a single body predicate, $prime/1$. 
The Herbrand Base, excluding the possible examples, is just the background knowledge. There are $o_p= 4$ literals which contain the $prime/1$ symbol out of a total $|HB|=16$ ($4$ $prime/1$, $6$ $even/1$, $6$ $odd/1$ literals). 
The cost of
$prime/1$ is thus $\text{predicates(r)} -\log{\frac{4}{16}} = 2$ bits (equation \ref{eq: predicate cost}).
\paragraph{Variables.} 
One single-argument predicate has only one variable arrangement, so $\text{vars}(r)=\log{1} =0$ bits (equation \ref{eq: variable cost}).
\paragraph{Theta values.}
Given the beta prior, the cost of sending the theta values $\theta_+=0.88, \theta_-=0.86$ using MML87 is $\text{theta}(h) = 6.5$  bits.
\paragraph{Coverage.} 
The program $\text{det}(h)$ entails all three observed examples in $E^+$ and $E^-$, therefore, $N_+=3$. Assuming a prior misclassification probability of $0.1$, the probability that $P(N_+=3) = 0.081$. The cost of stating $N_+$ is $\text{coverage}(h)=-\log{0.081}\approx 3.6$ bits.
\paragraph{Total hypothesis cost.} 
The total syntax cost $\text{syntax}(h)$ is the sum of $\text{structure}(h)$, $\text{predicates}(r)$ and $\text{vars}(h)$: $\text{syntax}(h)=0+2+0 =2$ bits.
The total hypothesis cost (equation \ref{eq:hyp length}) is therefore $C(h) = 2 + 6.5 + 3.6 =12.1$ bits.
\subsection*{Examples Cost}
We now describe communicating the examples under $h$. 
\paragraph{Atoms.} 
The deterministic hypothesis $\text{det}(h)$ entails all three training examples, and $4$ out of the $10$ possible examples $\{f(0), f(1), f(2), ... f(9)\}$.
The probability of choosing the $3$ literals in $E$ from the $4$ possible examples is $\binom{4}{3}^{-1}=0.25$, so $\text{atoms}(E|h)=-\log{0.25}=2$ bits.
\paragraph{Labels.}
The hypothesis $h$ classifies two examples correctly and one incorrectly under the $\theta_+$ probability model. The probability of making these errors is:
$0.88^2 \times (1-0.88) = 0.093$ so thus
$\text{labels}(E|h) = -\log{0.093} = 3.4$ bits.
\paragraph{Total cost.}
The total example length is therefore $2 + 3.4 = 5.4$ bits.
\subsection{Total Length}
The total length (equation \ref{eq:full cost}) is $12.1 + 5.4 = 17.5$ bits.
We search for the $\text{det}(h)$ which minimises the total length.

\section{Algorithms}
To test claims 1-5,  we need algorithms that (i) compare exact MML and MDL hypotheses and (ii) efficiently learn hypotheses with good MML and MDL scores. Our contribution in this section is two such algorithms: \textsc{RandomLearner} and \textsc{ApproxLearner}. 
\subsection{\textsc{RandomLearner}}
As the exact MML score is difficult to optimise, this learner calculates exact MML and MDL scores for a large number of random hypotheses, selecting hypotheses with the smallest MML and MDL scores. 
We generate $T$ random hypotheses for each task. The method of generating the random hypotheses follows. For each size $1-Max\_Size$, we generate a maximum of $I$ deterministic rules of this size which conform to the syntactic bias given the task. Rules are constrained to only contain at most $v\_max$ variables. 
From these rules, we generate a random, deterministic program, $\text{det}(h)$. 
We choose a random number $c$ between 1 and $Max\_Clauses$, and randomly select $c$ rules from the set of all generated rules. This random set of rules is a random program.
We repeat this $T$ times to generate all random programs.
We convert all programs $\text{det}(h)$ to probabilistic hypotheses $h$, scoring each as described.
Specific values for $T,\ Max\_Size,\ I,\ v\_max,\  c,$ and $Max\_Clauses$ may be found in Section 8.1.
\subsection{\textsc{ApproxLearner}}
This learner searches for hypotheses which minimise a user-specified cost function. In this paper, this cost is either C-MDL or an approximation of MML. The  \textsc{ApproxLearner} learning procedure is similar to the inductive ILP system \textsc{MaxSynth}, which minimises the C-MDL cost function \cite{hocquette2024learning}. 
Both procedures (i) generate individual rules, (ii) test and cache the size and coverage of each rule on training data, and (iii) find good combinations of these rules according to the cost.
As in \textsc{RandomLearner}, we assess a deterministic rule or combination of rules by treating it as $\text{det}(h)$ and converting to $h$ (definition \ref{def:hypothesis}) for scoring.
Pseudocode for this procedure is shown in Algorithm \ref{alg:approxlearner alg}. We describe these stages next.
\begin{algorithm}[h]
\small
{
\begin{myalgorithm}[]
def ApproxLearner(bk, pos, neg, max_size):
  bst_sol, bst_score, size = {}, $\infty$, 1
  poss = make_poss_exs(bk,pos,neg)
  seen_hyps, cons = {}, {}
  while size $\leq$ max_size:
    h = generate(cons, size)
    if h == UNSAT:
        size += 1
        continue
    tp,fn,tn,fp,gen = test(pos,neg,bk,poss,h)
    h_score = score(h,tp,fp,tn,fn,gen)
    if h_score < bst_score:
      bst_sol, bst_score = h, h_score
    seen_hyps += h
    comb = combine(seen_hyps, bst_score)
    if comb != UNSAT:
        bst_sol = comb
        test_res = test(pos,neg,bk,poss,comb)
        tp,fn,tn,fp,gen = test_res
        bst_score=score(comb,tp,fp,tn,fn,gen)
    cons += constrain(h,fp,fn)
  return bst_sol
\end{myalgorithm}
\caption{\textsc{ApproxLearner} algorithm}
\label{alg:approxlearner alg}
}
\end{algorithm}

\textbf{Constraints: } The main difference between \textsc{ApproxLearner} and \textsc{MaxSynth} is step (i). When generating rules, \textsc{MaxSynth} uses C-MDL-specific constraints to efficiently prune rules from the rule space that cannot be in a C-MDL optimal hypothesis. 
This pruning vastly shrinks the search space and enables \textsc{MaxSynth} to learn globally optimal hypotheses efficiently. 
Conversely, \textsc{ApproxLearner} uses weaker, cost function independent constraints (such as pruning generalisations of inconsistent rules) and is much less efficient.
In the absence of MML-specific constraints, we use the \textsc{ApproxLearner} procedure to compare MML and C-MDL costs fairly, leaving MML-specific constraints to future work.

\textbf{Search: } To implement the search for good rule combinations in step (iii), both \textsc{MaxSynth} and \textsc{ApproxLearner} use constraint solvers, specialised tools which efficiently search large combinatorial spaces given a cost to optimise. \textsc{ApproxLearner} uses the CP-SAT \cite{cpsatlp} constraint solver.
CP-SAT requires a linear cost to perform an effective search. The C-MDL cost is already linear and is modelled directly in CP-SAT. The MML cost is not linear, so we use a linear approximation.

\textbf{Linear MML Approximation: } We cache the values of 
all message costs that behave linearly. For all other terms that compose the $C(h,E)$ sum (equation \ref{eq:full cost}), we select five breakpoints and linearly interpolate between each to determine a piecewise linear (PL) approximation. We provide the PL approximations, the cached 
rule encoding values, and rule statistics to the CP-SAT solver so that it may implement the search procedure. 
 
\subsubsection{Herbrand Base Calculations}
For efficiency, in the \textsc{RandomLearner} experiments, the Herbrand Base is calculated once from the original trial and reused for any trial variation (altering training example size, noise level, or example balance). In the \textsc{ApproxLearner} experiments (\textbf{Q5} below), however, the Herbrand Base is recalculated for all trials. 

\section{Experiments}
We now test our claims.
Unless otherwise mentioned, all experiments use \textsc{RandomLearner}.
We first try to answer the question:
\begin{enumerate}
\item[\textbf{Q1a}] Does MML perform similarly to C-MDL on general tasks?
\end{enumerate}
We answer \textbf{Q1a} by comparing the testing accuracy of MML and C-MDL programs on 30 diverse datasets.

To test our claim that our MML method performs similarly to C-MDL on noisy data, we try to answer the question:
\begin{enumerate}
\item[\textbf{Q1b}] Can MML handle similar amounts of noise as C-MDL?
\end{enumerate}
We answer \textbf{Q1b} by introducing different amounts (10\%, 20\%, … 50\%) of random noise into each of the 30 datasets and comparing the testing accuracies of the MML and C-MDL selected programs.
We also run MML vs C-MDL on the real-world noisy \textit{alzheimer} datasets. We introduce noise by randomly selecting a subset of training examples, which we randomly reassign as positive or negative.

To test our claim that our MML approach can learn from unbalanced examples, we try to answer the question:
\begin{enumerate}
\item[\textbf{Q2}] Does MML outperform C-MDL when learning from different example proportions?
\end{enumerate}
We answer \textbf{Q2} by comparing the testing accuracy of MML and C-MDL programs on datasets with different proportions of examples, ranging from 10:0 (100\% positive examples), 8:2, 6:4, …, 0:10 (100\% negative examples). We randomly generate these datasets from those used in \textbf{Q1a}. We restrict dataset sizes to 20 and 50 to control for the effect of the number of training examples.

To test our claim that our MML approach is more data efficient than C-MDL, we try to answer the question:
\begin{enumerate}
\item[\textbf{Q3}] Can MML learn from fewer training examples than C-MDL?
\end{enumerate}
We answer \textbf{Q3} by comparing the testing accuracy of MML and C-MDL programs on training datasets of eight different training example sizes: 1, 5, 10, 20, 50, 100, 200, and 500. 
We build these datasets by random sampling from the original trial dataset.

To test our claim that the explicit preference for more general programs improves the data efficiency of our MML approach, we try to answer the question:
\begin{enumerate}
\item[\textbf{Q4}] Do generality priors allow MML to learn from fewer training examples?
\end{enumerate}
We answer \textbf{Q4} by comparing the testing accuracy of our standard MML procedure (with generality predicate priors) against MML with uniform predicate priors on the same datasets as \textbf{Q2}.

\begin{figure*}[t]
\centering
\includegraphics[width=1.0\textwidth]{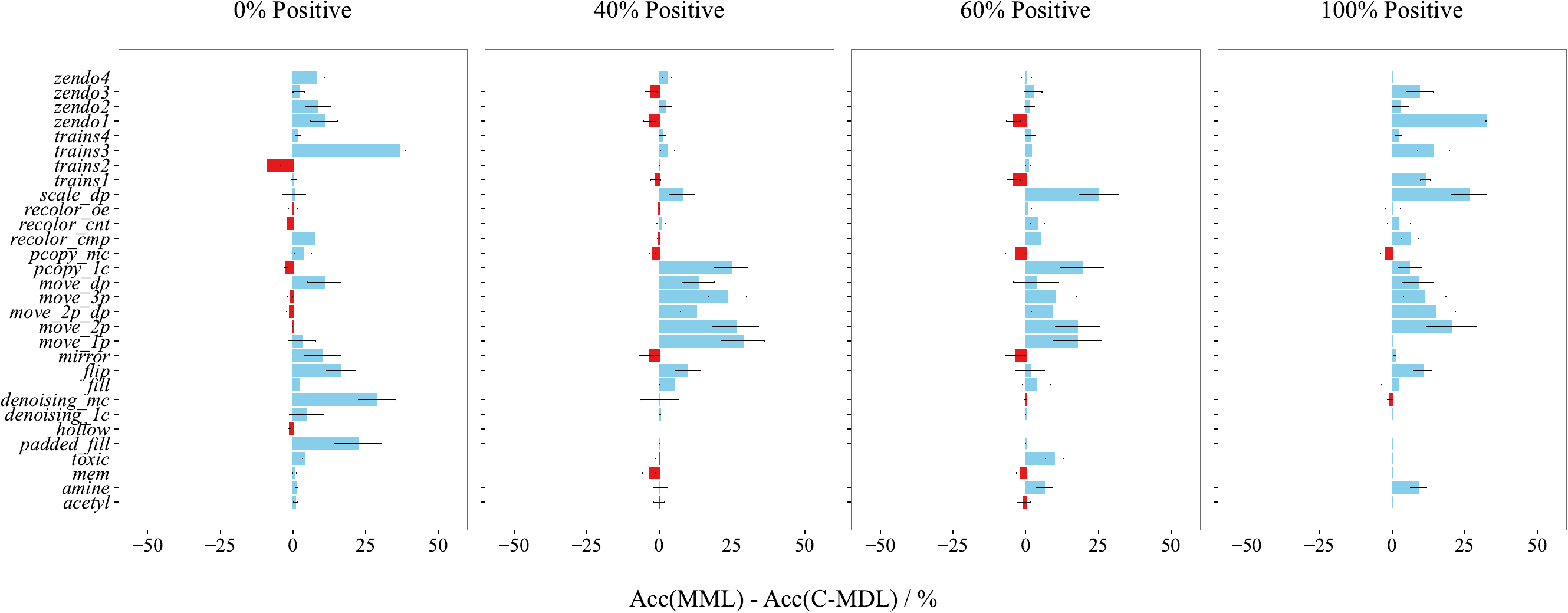} 
\caption{Balanced MML and C-MDL program accuracy differences when changing the proportion of positive examples for training datasets of size 20. Blanks indicate proportions not possible with 20 examples. Error bars represent standard error.}
\label{fig: example balance}
\end{figure*} 

To test our claim that an approximation to the MML cost function learns from unbalanced examples, we try to answer the question:
\begin{enumerate}
\item[\textbf{Q5}] Can a piecewise approximation to the MML cost enable learning from unbalanced examples?
\end{enumerate}
We answer \textbf{Q5} by comparing C-MDL and MML on the noiseless datasets in \textbf{Q1} and on modifications of these datasets containing only the positive or negative examples using \textsc{ApproxLearner}.
\subsection{Experimental Choices}
We make the following experimental choices.

\textbf{Theta Priors} 
The expected $\theta_+$ and $\theta_-$ values are $e=\frac{\alpha}{\alpha+\beta}$. The probability of misclassification error $r$, used in $C_{cov}(H)$ calculation is $r=1-e$. For noiseless tasks, we assume $r$ is near zero: $\alpha=1000,000$, $\beta=1$. For noisy tasks, we assume weaker, but still discriminatory programs: $\alpha=5000$, $\beta=1$. 
The beta priors chosen ensure $\theta_+, \theta_->0.5$.

\textbf{\textsc{RandomLearner} Parameters}
We generate $10,000$ random programs for each task.
Each rule has a maximum rule size of $4$ and at most $6$ variables. 
For each size, the maximum number of rules generated is $10,000$. 
Each random program contains at most $5$ rules.

\textbf{\textsc{ApproxLearner} Parameters}
We allow each CP-SAT call 180 seconds to find a good rule combination. 
We use a timeout of 1000 seconds for each trial. The best program found before timeout is returned. The score of the empty program provides the starting point for the search.

\textbf{Setup} We run all \textsc{RandomLearner} experiments on an M2 Macbook Air (16GB RAM, 8 core CPU) and \textsc{ApproxLearner} programs on a server (36 core CPU, 125GB RAM). We repeat all experiments 10 times and calculate the mean and standard error. We measure balanced testing accuracy for each task to account for imbalanced datasets. All significance claims are determined by Benjamini-Hochberg corrected Wilcoxon signed-rank tests at an overall significance level of 0.05. We make all comparisons between tasks, not trials within a task, to satisfy the test's independence assumptions.

\subsubsection{Datasets}
We test our claims on the following datasets:

\textbf{Alzheimer}. Real-world noisy datasets, learning attributes of drug molecules for alzheimers treatment \cite{King92}.

\textbf{Zendo}. An inductive game where the goal is to determine an underlying true rule \cite{discopopper}.

\textbf{1D-ARC}. A one-dimensional adaptation \cite{xu2023llms} of the ARC-AGI dataset \cite{chollet:2019}.

\textbf{Trains}. Synthetic datasets whose goal is to determine a rule which differentiates eastbound from westbound trains \cite{larson1977inductive}.
\subsection{Results}
The experimental results follow.
\subsubsection{Q1a: Overall Performance: \textsc{RandomLearner}} 
On the 30 standard tasks, we find no statistically significant difference between MML and C-MDL. We hypothesise that this is because most datasets are balanced and/or moderately sized, qualities which allow C-MDL and standard ILP systems to perform well. These results support \textbf{Claim 1}. 
\subsubsection{Q1b: Noise Tolerance: \textsc{RandomLearner}}
On datasets with 10-50\% added noise or the standard \textit{alzheimers} datasets, we find no statistically significant difference between MML and C-MDL. We hypothesise that MML's performance in these low-signal domains is due to a prior that is overly optimistic, and would be improved by one which is more carefully selected. 
These results support \textbf{Claim 1}. 
\subsubsection{Q2: Example Balance: \textsc{RandomLearner}}
For training sets of size 20 and 50, the mean pairwise balanced accuracy difference consistently favours MML across all 12 tested size-example proportion datasets. The advantage is statistically significant ($p<0.05$) in 11/12 conditions. 
Figure \ref{fig: example balance} illustrates this trend for the size 20 datasets. 
One reason why MML outperforms C-MDL is because C-MDL struggles when learning from 100\% positive examples. In this setup, a program that classifies all examples as true is C-MDL optimal (it makes no errors and is minimal in size).
By contrast, the MML likelihood (Equation \ref{eq: second part}) penalises overly general programs.
A program that entails all possible examples results in the cost of $\text{atoms}(E|h)$ as $\log\binom{|T|}{tp+fp}$ which, when $|T| >> tp+fp$, is very large. 
MML prefers a program which entails only the seen examples (and makes no/few errors), as this cost is instead $\log\binom{tp+fp}{tp+fp} = 0$. 
These results support \textbf{Claim 2}.
\subsubsection{Q3: Data Efficiency: \textsc{RandomLearner}} 
MML achieves a statistically significant improvement in the mean pairwise accuracy difference over C-MDL at all eight tested training sizes. On 5 out of the 8 tested sizes, MML results in more than a 4\% average improvement over all tasks. The improvement is largest for datasets of size 1, yielding an improvement of $6.8\pm2.4\%$. We hypothesise this improvement is due to our generality prior coupled with the more informative likelihood. These results support \textbf{Claim 3}. 
\subsubsection{Q4: Generality Priors: \textsc{RandomLearner}} 
The standard MML procedure with generality priors achieves a statistically significant ($p<0.05$) improvement of $6.6\pm2.6\%$ in mean pairwise
accuracy difference over MML with uniform priors when learning from a single example. We find no significant difference at any larger training size. These results support \textbf{Claim 4}.
\subsubsection{Q5: Example Balance Tolerance and Overall Performance: \textsc{ApproxLearner}}
On datasets consisting of exclusively positive and exclusively negative examples, the approximate MML cost significantly ($p<0.01$) outperforms C-MDL by $9.2\pm2.7$\% and $25.9\pm4.0\%$ respectively. The maximum improvement was 50\% in both experiments: the maximum possible. 
The maximum losses were much smaller, $-4.8\%$ and $-0.0\%$ for $100\%$ and $0\%$ positive examples, respectively.
Experiments on the standard noiseless tasks reveal no significant difference between MML and C-MDL. This result is despite the difficulty of optimising the more complex MML cost function. These results support \textbf{Claim 5}

\section{Conclusions and Limitations}
We derived an ILP cost function that is data-efficient, insensitive to example imbalance, and noise-tolerant. 
Our cost function enables learning from exclusively positive examples, a common limitation for many ILP systems \cite{popper,hocquette2024learning}.
Empirically, our cost function outperforms the that of the state-of-the-art MDL system \textsc{MaxSynth} on data efficiency and example imbalance, and matches its performance on noisy data.
Our prior preference for more general predicates results in significantly increased data efficiency relative to a uniform prior when learning from a single example.
\subsection*{Limitations}
Unlike \textsc{MaxSynth}, our system does not guarantee optimal programs. 
Future work should identify better constraints for \textsc{ApproxLearner} to  enable efficient optimal search for MML-based systems. 
The described approach has only been demonstrated for finite domains, but may be extended to infinite domains by constructing a truncated Herbrand Base from only the ground terms observed. 
Alternative priors should also be explored, such as extending the presented predicate generality priors to incorporate domain knowledge. 
The beta prior requires user specification, which is not needed for MDL. 
One should instead determine robust noisy default beta priors by grid search over many datasets, which should improve performance, particularly in high-noise domains. 


\end{document}